\documentclass{article}

\usepackage{microtype}
\usepackage{graphicx}
\usepackage{subfigure}
\usepackage{booktabs} 
\usepackage{amsmath}
\usepackage{amssymb}
\usepackage{multirow}
\usepackage{titlesec}
\titleformat{\paragraph}
{\normalfont\normalsize\bfseries}{\theparagraph}{1em}{}
\titlespacing*{\paragraph}
{0pt}{3.25ex plus 1ex minus .2ex}{1.5ex plus .2ex}
\usepackage{hyperref}

\usepackage[accepted]{icml2019}

\icmltitlerunning{Text Classification and Clustering with Annealing Soft Nearest Neighbor Loss}

\begin{document}

\twocolumn[
\icmltitle{
Text Classification and Clustering with Annealing Soft Nearest Neighbor Loss
}
% \icmltitle{Disentangling Natural Language Representations with Annealing Soft Nearest Neighbour Loss}




\begin{icmlauthorlist}
\icmlauthor{Abien Fred Agarap}{dlsu}
\end{icmlauthorlist}

\icmlaffiliation{dlsu}{CSC715M Natural Language Processing Final Project\\College of Computer Studies, De La Salle University, Manila, Philippines}

\icmlcorrespondingauthor{Abien Fred Agarap}{abien\_agarap@dlsu.edu.ph}

\icmlkeywords{Machine Learning, ICML}

\vskip 0.3in
]



\printAffiliationsAndNotice{}

\begin{abstract}
We define disentanglement as how far class-different data points from each other are, relative to the distances among class-similar data points. When maximizing disentanglement during representation learning, we obtain a transformed feature representation where the class memberships of the data points are preserved. If the class memberships of the data points are preserved, we would have a feature representation space in which a nearest neighbour classifier or a clustering algorithm would perform well. We take advantage of this method to learn better natural language representation, and employ it on text classification and text clustering tasks. Through disentanglement, we obtain text representations with better-defined clusters and improve text classification performance. Our approach had a test classification accuracy of as high as 90.11\% and test clustering accuracy of 88\% on the AG News dataset, outperforming our baseline models -- without any other training tricks or regularization.
\end{abstract}

\section{Introduction and Related Works}
Neural networks are automated solutions on machine learning tasks such as image or text, classification, language translation, and speech recognition among others. They act as function approximation models that compose a number of hidden layer representations, in the form of nonlinear function, to come up with better representations for downstream tasks \cite{bengio2013representation, rumelhart1985learning}. In this work, we focus on the usage of learned representations by a neural network on natural language processing tasks, particularly, on text classification and text clustering.\\
Natural language processing is a task in artificial intelligence that deals with analysis and synthesis of natural language data, and among its sub-tasks include but are not limited to classification, clustering, and synthesis of speech and text. Like all other types of data for tasks in artificial intelligence, natural language data are required to be represented in numerical form. However, it has been difficult to represent text data in a way that is convenient for computational models. In the early years, the most common paradigm of text representation is through the use of techniques such as n-Grams and Term Frequency-Inverse Document Frequency (or TF-IDF).
\subsection{n-Grams}
In n-Grams, we represent a text in a contiguous sequence of \textit{n} tokens, e.g. ``I like dancing in the rain'' will have a unigram ($n=1$) representation of each word acting as a token, i.e. [``I'', ``like'', ``dancing'', ``in'', ``the'', ``rain''] -- and it follows that if $n=2$, each token will consist of pairs of words from the given text sequence. This technique allows us to capture the context of words that are frequently used together, e.g. a bigram ($n=2$) of ``New York'' will imply the city or state instead of having it in unigrams (``New'' and ``York''). However, the drawbacks from this technique are the sparsity of its representations, and its inability to take into account the order of words in how they appear in a document, and finally, it suffers from the \textit{curse of dimensionality}.
\subsection{TF-IDF}
In TF-IDF, we represent a text by scoring the importance of a word with respect to a document and to the entire corpus, e.g. a word that occurs multiple times in a document but less times in the entire corpus will be deemed as important, and will have a high score, but a word that occurs multiple times in a document and in the entire corpus will be deemed as a common word, and thus will have a low score. Similar to n-Grams, TF-IDF does not take into account the order of words in a document. In addition, we cannot capture the context of meaning words in TF-IDF.
\subsection{Word Embeddings}
To solve the problem of sparsity and the inability to capture the contextual meaning of words, \citeauthor{mikolov2013distributed} introduced an efficient word embeddings representation. This representation learns word vector representations that capture semantic word relationships. In this technique, each word is represented as a weight vector from a neural network that was designed to predict the neighboring tokens of a word, i.e. the input is the word \textit{w} at time \textit{t} and the outputs are the neighboring words of $w_{t}$. Since then, there have been a number of word embeddings proposed such as GloVe \cite{pennington2014glove} and fastText \cite{mikolov2018advances}.
\subsection{Sentence Embeddings}
Word embeddings are a result of an unsupervised training on large corpora, and they have been the basis of numerous advancements in natural language processing \cite{devlin2018bert, joulin2016fasttext, mikolov2013distributed, pennington2014glove, peters2018deep} during the past decade. While they have been quite useful to capture the contextual meaning of words and to represent text data in downstream tasks, it was not clear whether they can capture the meaning of a full sentence, i.e. to embed a full sentence instead of just words. Among the early attempts to provide sentence embeddings was to get the vector representation for each word in a sentence, and use their average as the embeddings for the full sentence \cite{arora2016simple}. Despite being a strong baseline for sentence embeddings, the efficacy of this method in capturing relationships among words and phrases in a single vector warrants further investigations. One of the successful attempts to obtain embeddings for full sentences was InferSent \cite{conneau2017supervised}, which also uses the word embeddings approach for representing each token in a text sequence, then feeds it to a bidirectional recurrent neural network with long-short term memory \cite{hochreiter1997long} and max pooling architecture. The resulting learned representation by this network is then taken as the full sentence embedding.
\subsection{Our contributions}
In our work, we use both methods for computing sentence embeddings, i.e. averaged word embeddings from \cite{honnibal2017spacy}, and InferSent embeddings \cite{conneau2017supervised}, for text representation in our classification and clustering tasks. Doing so will provide an additional literature to the comparison of use and effectiveness of the aforementioned techniques for computing sentence embeddings. Using these embeddings, we introduce the idea of \textit{disentangling} internal representations of a neural network to further improve the performance on downstream NLP tasks. In this context, \textit{disentangling} means we transform the representation to preserve the class membership of the data points. To the best of our knowledge, this is the first work on disentangling text representations. 

\section{Disentangling Text Representations}
We consider the problem of text classification and text clustering using averaged word embeddings and sentence embeddings as the representation scheme. To improve the performance on the classification task, we transform the feature representations learned in the hidden layers of a neural network to preserve the class memberships of the data points. In this transformation, we maximize the distances among class-different data points. In doing so, we also obtain a more clustering-friendly representation since in this transformed space, the inter-cluster variance will be maximized.

\subsection{Feed-Forward Neural Network}
The feed-forward neural network (also colloquially known as deep neural network) is the quintessential deep learning model that learns to approximation the function mapping between the input and output of a dataset, i.e. $y \approx f\left( \Vec{x}; \theta \right)$. Its parameters $\theta$ are optimized to learn the best approximation of the input targets, which may be a class label or a real value.
\begin{align}\label{eq:hidden_layers}
    f(\Vec{x}) = f^{(n)}\left( f^{(\ldots)} f^{(1)}(\Vec{x}) \right)
\end{align}
\begin{align}\label{eq:cross_entropy}
    \ell_{ce}(y, f(x)) = -\sum_{i} y_{i} \log \left[ f(x_{i}) \right]
\end{align}
In order to do this, the model composes a number of nonlinear functions in the form of hidden layers, each of which learns a representation of the input features (see Eq. \ref{eq:hidden_layers}). Afterwards, to optimize the model it measures the similarity between the approximation and the input targets by using an error function such as cross entropy (see Eq. \ref{eq:cross_entropy}).\\
For our experiments, we use two neural network architectures. The first network we used was for the text representation using averaged word embeddings as the sentence embeddings, which had two hidden layers with 500 neurons each. The second network we used was the for the text representation using InferSent sentence embeddings, which had one hidden layer with 500 neurons. The hidden layers were initialized with Kaiming initializer \cite{he2015delving} and had ReLU \cite{nair2010rectified} as the nonlinearity function, while the output layer was initialized with Xavier initializer and used the softmax nonlinearity function.\\
We transform the hidden layer representations of a neural network to maximize the distances among class-different data points. We accomplish this by using the soft nearest neighbor loss\cite{agarap2020improving, frosst2019analyzing, salakhutdinov2007learning}, which we will discuss later in this chapter.

\subsection{Convolutional Neural Network}
The convolutional neural network (or CNN) is a neural network variant that uses the convolution operator as the feature extractor in its hidden layers \cite{lecun1998gradient}. Similar to feed-forward neural networks, they also compose hidden layer representations for a downstream task. However, in convolutional neural networks, they use the hierarchical nature of data, and assemble more complex patterns using smaller and simpler patterns which are computed using the convolution operator. We used an architecture with one 1D convolutional layer followed by three fully connected layers with 2048, 1024, and 512 neurons. All the hidden layers were initialized with Kaiming initializer and used the ReLU nonlinearity function, while the output layer was initialized with Xavier initializer and used the softmax nonlinearity function.\\
Similar to our feed-forward network, we transform the hidden layer representations of our convolutional neural network to disentangle the feature representations using the soft nearest neighbor loss.

\subsection{Autoencoder}
An autoencoder \cite{bourlard1988auto, hinton1994autoencoders, hinton2006reducing, vincent2008extracting, vincent2010stacked} is a neural network that aims to find the function mapping for the features $x$ to itself through the use of an encoder function $h = enc(x)$ that learns the latent code representation for the features, then a decoder function that reconstructs the original features from the latent code representation $r = dec(h)$. The latent code representation has a lower dimensionality than the original feature representation, in which the most salient features of the data are learned. Intuitively, the encoder and the decoder layers can be thought of as individual neural networks that are stacked together to form an autoencoder.\\
To learn the reconstruction task, it minimizes a loss function $\mathcal{L}(x, dec(enc(x)))$, where $\mathcal{L}$ is a function penalizing the decoder output $dec(enc(x))$ for being dissimilar from the original features $x$. Typically, this reconstruction loss is the Mean Squared Error (MSE) $\frac{1}{n} \sum_{i=1}^{n}\|dec(enc(x_{i})) - x_{i}\|_{2}^{2}$. Then, like other neural networks, it is usually trained using a gradient-based method aided with the backpropagation of errors \cite{rumelhart1985learning}.
\begin{align}\label{eq:sigmoid_recon_loss}
    \ell_{rec}(x, r) = \dfrac{1}{n} \sum_{i=1}^{n} -x_{i}\log(r_{i}) + (1 - x_{i})\log(1 - r_{i})
\end{align}
The reconstruction task of an autoencoder has a by-product of learning good feature representations, and we take advantage of this for clustering, particularly, we use the latent code representation $z = enc(x)$ as the input features for the clustering algorithm. Our autoencoder had a $d-500-500-2000-z-2000-500-500-\hat{d}$ architecture, where $d$ and $\hat{d}$ are the dimensionality of the features, and $z$ is the dimensionality of the latent code, which we set to 128 in all our experiments. We use the binary cross entropy (see Eq. \ref{eq:sigmoid_recon_loss}) as the reconstruction loss for our experiments. All the hidden layers were initialized with Kaiming initializer and used ReLU as the nonlinearity function, while the encoder and decoder output layers were initialized with Xavier initializer and used the logistic nonlinearity function.
\begin{figure}[htb]
    \centering
    \includegraphics[width=0.90\linewidth]{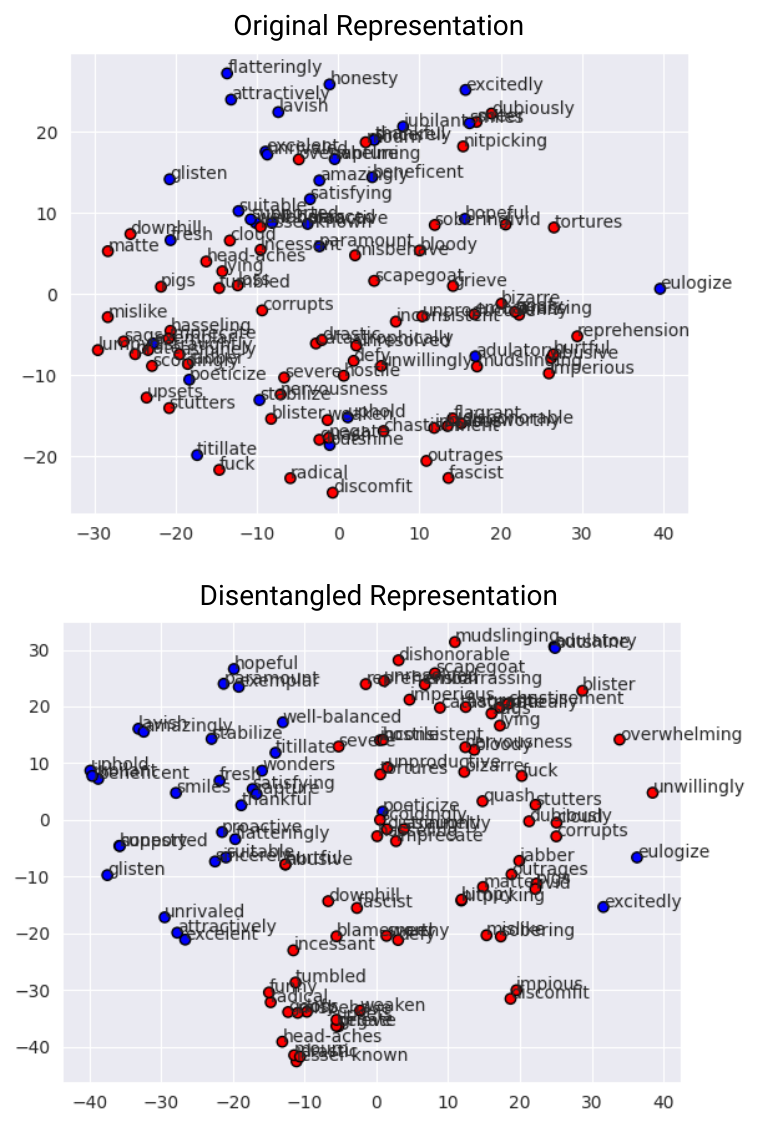}
    \caption{A subset of 100 word vectors from an opinion lexicon \cite{hu2004mining}. We trained an autoencoder network for 50 epochs to learn a disentangled latent code representation for an opinion lexicon. We can see the disentangled word vectors for positive and negative words denoted by colors blue and red respectively. In the transformed representation, we can see a better clustering of the positive and negative words respectively. This figure is best viewed in color.}
    \label{fig:opinion_lexicon}
\end{figure}
\subsection{Soft Nearest Neighbor Loss}
In the context of our work, we define \textit{disentanglement} as how close pairs of class-similar feature representations are, relative to pairs of class-different feature representations. \citeauthor{frosst2019analyzing} used the same term in the same context. A low entanglement value implies that feature representations from the same class are closer together than they are to feature representations from different classes. To measure the entanglement of feature representations, \citeauthor{frosst2019analyzing} expanded the non-linear neighborhood component analysis (NCA) \cite{salakhutdinov2007learning} objective by introducing the temperature factor $T$, and called this modified objective the \textit{soft nearest neighbor loss}.\\
They defined the \textit{soft nearest neighbor loss} as the non-linear NCA at temperature $T$, for a batch of $b$ samples $(x, y)$,
\begin{equation}\label{eq:snnl}
    \ell_{snn}(x, y, T) = -\dfrac{1}{b} \sum_{i \in 1 \dots b} \log \left( \dfrac{\sum\limits_{\substack{j \in 1 \dots b \\ j \neq i \\ y_{i} = y_{j}}} \exp\left(-\dfrac{d_{ij}}{T}\right) }{\sum\limits_{\substack{k \in 1 \dots b \\ k \neq i}} \exp\left(-\dfrac{d_{ik}}{T}\right)} \right)
\end{equation}
where $d$ is a distance metric on either raw input features or learned hidden layer representations $x$ of a neural network, and $T$ is the temperature factor that is directly proportional to the distances among the data points. We use the pairwise cosine distance (see Eq. \ref{eq:cosine_distance}) as our distance metric on all our experiments for more stable computations.
\begin{align}\label{eq:cosine_distance}
    d_{ij} = 1 - \dfrac{\sum_{n}^{N} x_{i} x_{j}}{\sqrt{\sum_{n}^{N} x_{i}^{2}} \sqrt{\sum_{n}^{N} x_{j}^{2}}}
\end{align}
Similarly, we employ the annealing temperature (see Eq. \ref{eq:annealing_temp}) proposed by \citeauthor{agarap2020improving} for more stable computations.
\begin{equation}\label{eq:annealing_temp}
    T = \dfrac{1}{(\eta + i)^{\gamma}}
\end{equation}
where $i$ is the current training epoch index, and we set $\eta = 1$ and $\gamma = 0.55$ for our experiments, similar to \citeauthor{neelakantan2015adding}.

\citeauthor{frosst2019analyzing} described the temperature factor as a means to control the relative importance of the distances between pairs of points, that is, the loss is dominated by small distances when using a low temperature.
We can describe the soft nearest neighbor loss as the negative log probability of sampling a neighboring data point $j$ from the same class as $i$ in a mini-batch $b$, which is similar to the probabilistic sampling by \citeauthor{goldberger2005neighbourhood}. A low soft nearest neighbor loss value is implies a low entanglement (or high disentanglement).\\
In Figure \ref{fig:opinion_lexicon}, we show a disentangled latent code representation for the word vectors from an opinion lexicon by \cite{hu2004mining}. In the aforementioned figure, we can see a better separation of the word vectors from the two classes.

\section{Downstream Tasks on Disentangled Representations}
We demonstrate the effectiveness of our approach by conducting experiments on a benchmark dataset, and lay down the classification and clustering performance of our baseline and experimental models.

\subsection{Dataset}
Due to time and computational restraints, we were only able to run experiments on one dataset. In future works, we intend to use more datasets. In the current study, we use the AG News dataset, a text classification dataset that is comprised of four classes \cite{zhang2015character}. Despite the training set having a 119,843 documents, we only use a subset of it in select experiments due to time and computational resource constraints. However, we still used the full test set of 7,600 documents for our evaluations. We removed the English stop words, the words that have less than 3 characters, and the non-alphanumeric characters. Then, we normalized the texts into lowercase.\\
We computed the sentence embeddings before training to save time and computation resources. For the first set of sentence embeddings, we computed the average GloVe \cite{pennington2014glove} word embeddings for each example using SpaCy \cite{honnibal2017spacy}, while for the second set of sentence embeddings, we computed the InferSent embeddings for each example.\\
Our first set of sentence embeddings had 300 dimensions for each example, while the second set of embeddings had 4096 dimensions for each example. The large gap between the dimensionalities of the sentence embeddings we used was due to the use of bidirectional RNN-LSTM in InferSent.

\subsection{Experimental Setup}
We used a computing machine with Intel Core i5-6300 HQ (2.60 GHz) with 16 GB DDR3 RAM and Nvidia GTX 960M 4GB GDDR5 GPU. We ran our computations for each of our experiments for five times, each of which used a pseudorandom seed for reproducibility. The seeds we used were as follows: 42, 1234, 73, 1024, and 31415926.\\
For our experiments where we used the averaged word embeddings as the sentence embeddings, we only used a pseudorandomly picked subset with 20,000 examples while we still used the full test with 7,600 examples for evaluation. As for our experiments where we used the InferSent embeddings as the sentence embeddings, we used the full training set with 119,843 examples and the full test with 7,600 examples for evaluation.

\subsection{Text Classification}
We trained feed-forward neural networks and and convolutional neural networks using a composite loss (see Eq. \ref{eq:composite_loss_classification}) of the cross entropy as the classification loss, and the soft nearest neighbor loss as the regularizer.
\begin{align}\label{eq:composite_loss_classification}
    \mathcal{L}(f, x, y) = \ell_{ce}\left(y, f(x)\right) + \alpha \cdot \sum_{b \in B} \ell_{snn} \left(f^{(b)}(x), y\right)
\end{align}
where $b$ is the index of the hidden layer of the neural network. We did not perform any hyperparameter turning or any training tricks since we only want to show that we can use disentanglement for some natural language processing tasks. Our neural networks were trained using Adam \cite{kingma2014adam} optimization with a learning rate of $1\times10^{-3}$ on a mini-batch size of 256 for 30 epochs for the feed-forward neural networks and 50 epochs for the convolutional neural networks. We used an $\alpha$ value of 100 for disentanglement and -100 for entanglement. The rationale for entanglement is based on the findings by \citeauthor{frosst2019analyzing}, where they showed that entangling the representations in the hidden layer of a network results to an even better disentanglement in the classification layer.\\
\begin{table}[htb!]
    \centering
    \resizebox{\linewidth}{!}{%
    \begin{tabular}{| *{9}{c|}}
        \hline
        \multirow{3}{*}{Model}  &   \multicolumn{4}{c|}{SpaCy Embeddings}   &   \multicolumn{4}{c|}{InferSent Embeddings}   \\
        \cline{2-9}
                                &   \multicolumn{2}{c|}{ACC}    &   \multicolumn{2}{c|}{F1}     &   \multicolumn{2}{c|}{ACC}    &   \multicolumn{2}{c|}{F1} \\
        \cline{2-9}
        &   AVG &   MAX &   AVG &   MAX     &   AVG &   MAX &   AVG &   MAX \\
        \hline
        Baseline	&   87.31\% &   88.18\%	&   0.8732	&   0.8818   &  89.34\% &   89.74\% &	0.8934	&   0.8974  \\
        \textbf{SNNL 100}	&   \textbf{88.02\%} &   \textbf{88.22\%}	&   \textbf{0.8802}	&   \textbf{0.8822}   &     89.58\% &   89.68\%	&   0.8958	&   0.8968  \\
        \textbf{SNNL -100}   &   81.71\%  &   82.71\%  &   0.8109  &   0.8271  &   \textbf{89.60\%}	&   \textbf{90.11\%}	&   \textbf{0.8960}	&   \textbf{0.9011} \\
        \hline
    \end{tabular}}
    \caption{Using a composite loss, which minimizes cross entropy loss and optimizes entanglement through the soft nearest neighbor loss, has a marginal increase in test performance on the AG News dataset classification using a feed-forward neural network.}
    \label{tab:dnn_classification}
\end{table}
\begin{table}[htb!]
    \centering
    \resizebox{\linewidth}{!}{%
    \begin{tabular}{| *{9}{c|}}
        \hline
        \multirow{3}{*}{Model}  &   \multicolumn{4}{c|}{SpaCy Embeddings}   &   \multicolumn{4}{c|}{InferSent Embeddings}   \\
        \cline{2-9}
                                &   \multicolumn{2}{c|}{ACC}    &   \multicolumn{2}{c|}{F1}     &   \multicolumn{2}{c|}{ACC}    &   \multicolumn{2}{c|}{F1} \\
        \cline{2-9}
        &   AVG &   MAX &   AVG &   MAX     &   AVG &   MAX &   AVG &   MAX \\
        \hline
        Baseline	&   87.08\% &   87.42\%	&   0.8708	&   0.8742   &  86.92\% &   88.42\% &	0.8692	&   0.8842  \\
        \textbf{SNNL 100}	&   \textbf{87.19\%} &   \textbf{87.89\%}	&   \textbf{0.8719}	&   \textbf{0.8789}   &     \textbf{87.67\%} &   \textbf{88.74\%}	&   \textbf{0.8767}	&   \textbf{0.8874}  \\
        SNNL -100   &   49.98\%  &   75.82\%  &   0.4998  &   0.7582  &   25\%	&   25\%	&  0.25	&   0.25 \\
        \hline
    \end{tabular}}
    \caption{Using a composite loss, which minimizes cross entropy loss and optimizes entanglement through the soft nearest neighbor loss, has a marginal increase in test performance on the AG News dataset classification using a convolutional neural network.}
    \label{tab:cnn_classification}
\end{table}

In Tables \ref{tab:dnn_classification} and \ref{tab:cnn_classification}, we can see marginal improvement in the test classification performance of our experimental model. In Table \ref{tab:dnn_classification}, we can see the test classification performance of our feed-forward neural network on the averaged GloVe word embeddings as the sentence embeddings. Through disentanglement, our network performs better than the baseline model but performs worse through entanglement. We may attribute this to our use of relatively shallow neural networks, i.e. having two hidden layers at most, while the models in \cite{frosst2019analyzing} were much deeper, e.g. having five hidden layers at the minimum. We posit that entanglement or disentanglement works even better with deeper models. However, on the InferSent embeddings, we see a marginal increase in test classification performance when using entanglement. In Table \ref{tab:cnn_classification}, we see a consistent performance by our convolutional neural network when disentangling hidden layer representations, outperforming the baseline and entanglement models.

\subsection{Text Clustering}
We trained an autoencoder network using a composite loss (see Eq. \ref{eq:composite_loss_reconstruction}) of the binary cross entropy as the reconstruction loss, and the soft nearest neighbor loss as the regularizer.
\begin{align}\label{eq:composite_loss_reconstruction}
    \mathcal{L}(f, x, y) = \ell_{rec}\left(x, f(x)\right) + \alpha \cdot \sum_{b \in B} \ell_{snn} \left(f^{(b)}(x), y\right)
\end{align}
Similar to our classification experiments, we did not perform any hyperparameter tuning or any training tricks. Our autoencoder was trained using Adam optimization with a learning rate of $1\times10^{-3}$ on a mini-batch size of 256 for 30 epochs. Unlike the classification experiments, we only used an $\alpha$ value of 100 due to time constraints. For autoencoding, we had two modes of disentanglement: the first is by disentangling only a part of the latent code, particularly, only 100 of its 128 units \`{a} la \cite{salakhutdinov2007learning}, and the second is by disentangling all the hidden layers of the autoencoder network.\\
Afterwards, we use the disentangled latent code, from these two modes of disentanglement, as the input features for our k-Means clustering algorithm \cite{lloyd1982least}.

We used the number of ground-truth categories in our dataset as the number of clusters. We used six different metrics to evaluate the clustering performance of our baseline and experimental models. The first three metrics fall under the category of internal criteria for clustering evaluation, which measure clustering quality. The second three metrics fall under the category of external criteria for clustering evaluation, which measure clustering membership.

\subsubsection{Internal Criteria}
In internal criteria for clustering evaluation, the clustering is subjected to optimizing the intra- and inter-cluster similarity. Specifically, to obtain a high intra-cluster similarity, and a low inter-cluster similarity. These optimization metrics are based on the cluster quality alone, without any external validation. We note that good scores in internal criteria metrics may not necessarily imply the effectiveness of the clustering model.

\paragraph{ Davies-Bouldin Index} (DBI) is the ratio between the intra-cluster distances and the inter-cluster distances \cite{davies1979cluster, halkidi2001clustering}. A low DBI denotes a good cluster separation. We compute it using Eq. \ref{db-index},
\begin{align}\label{db-index}
    DBI = \dfrac{1}{k} \sum_{i=1}^{k} \max_{i \neq j} R_{j}
\end{align}
\indent where $R$ is the mean similarity among clusters given by the following equation,
\begin{align}\label{average-similarity}
    R_{ij} = \dfrac{s_{i} + s_{j}}{d_{ij}}
\end{align}
\indent where $s$ is the cluster diameter which is the average distance between each point in the cluster and the cluster centroid, and $d_{ij}$ is the distance between centroids $i$ and $j$.

\paragraph{ Silhouette Score} (SIL) measures the similarity among examples in their own cluster compared to other clusters \cite{rousseeuw1987silhouettes}, and we compute it using Eq. \ref{silhouette-score},
\begin{align}\label{silhouette-score}
    SIL = \dfrac{b(i) - a(i)}{\max\left[a(i), b(i)\right]}
\end{align}
where $a(i)$ is the average within-cluster distances, and we compute it using the following equation,
\begin{align}\label{silhouette-score-a}
    a(i) = \dfrac{1}{|C_{i}| - 1} \sum_{j \in C_{i}, i \neq j} d(i, j)
\end{align}
where $C_{i}$ is the predicted cluster for point $i$, and $d(i, j)$ is the distance between points $i$ and $j$. Then, $b(i)$ is the average nearest-cluster distance, and we compute it using the following equation,
\ref{silhouette-score-b},
\begin{align}\label{silhouette-score-b}
    b(i) = \min_{k \neq i} \dfrac{1}{|C_{k}|} \sum_{j \in C_{k}} d(i, j)
\end{align}
Although any distance metric may be used, we used the Euclidean distance for our evaluations.

\paragraph{ Calinski-Harabasz Score} (CHS) is the ratio of intra-cluster dispersion to inter-cluster dispersion, and we compute it using Eq. \ref{ch-score}. A high CHS denotes good cluster separation \cite{calinski1974dendrite}.
\begin{align}\label{ch-score}
    CHS = \dfrac{Tr(B_{k})}{Tr(W_{k})} \times \dfrac{N - k}{k - 1}
\end{align}
where $B_{k}$ is the inter-cluster dispersion matrix given by the following,
\begin{align}\label{bk}
    B_{k} = \sum_{q} n_{q} (c_{q} - c)(c_{q} - c)^{T}
\end{align}
and $W_{k}$ is the intra-cluster dispersion matrix given by the following,
\begin{align}\label{wk}
    W_{k} = \sum_{q = 1}^{k}\sum_{x \in C_{q}} (x - c_{q})(x - c_{q})^{T}
\end{align}
where $N$ is the number of points in the data, $C_{q}$ contains the data of points in cluster $q$, $c_{q}$ is the center of cluster $q$, $c$ is the center of $C$, and $n_{q}$ is the number of data points in cluster $q$.

\subsubsection{External Criteria}
In external criteria for clustering evaluation, we use external information such as the class labels of a benchmark dataset for our validation. The external criteria metrics measure the clustering membership of the data points by using the class labels as the pseudo-cluster labels. Using the ground-truth labels as the pseudo-cluster labels is based on the \textit{cluster assumption} in the semi-supervised learning literature \cite{chapelle2009semi}.

\paragraph{ Normalized Mutual Information} (NMI) is the normalization of the mutual information (MI) score to transform its value to the $[0, 1]\in\mathbb{R}$ range, where 0 implies no mutual information while 1 implies perfect correlation. We compute it using the following,
\begin{equation}
    NMI = \dfrac{2I(y, c)}{[H(y) + H(c)]}
\end{equation}
where $y$ is the pseudo-cluster label, $c$ is the predicted cluster label, $H$ is the entropy, and $I$ is the mutual information between the pseudo-cluster labels and the predicted cluster labels.

\paragraph{ Adjusted Rand Index} (ARI) is the Rand Index (RI) adjusted for chance \cite{hubert1985comparing}. It measures the similarity between two clusterings by iterating through all pairs of points and counting pairs assigned to the same or different clusters according to the pseudo-cluster labels and the predicted cluster labels (see Eq. \ref{eq:ri}). 
\begin{equation}\label{eq:ri}
    RI = \dfrac{TP + TN}{TP + FP + FN + TN}
\end{equation}
where $TP$ is the true positive, $TN$ is the true negative, $FP$ is the false positive, and $FN$ is the false negative. We then compute ARI from RI through the following,
\begin{equation}\label{eq:ari}
    ARI = \dfrac{RI - \mathbb{E}[RI]}{\max(RI) - \mathbb{E}[RI]}
\end{equation}
ARI values are in $[0, 1]\in\mathbb{R}$ range, where 0 implies random labelling independently of the number of clusters, while 1 implies the clusterings are identical up to a permutation.

\paragraph{ Clustering Accuracy} (ACC) is the best match between the pseudo-cluster labels and the predicted clusters \cite{yang2010image}.
\begin{equation}
    ACC = \max_{m} \dfrac{\sum\limits_{i=1}^{n}1\{l_{i}=m\left(c_{i}\right)\}}{n},
\end{equation}

where $l_{i}$ is the pseudo-cluster label, $c_{i}$ is the predicted cluster label, and $m$ ranges over all possible one-to-one mappings between the labels and the clusters.

\begin{table*}[h]
    \centering
    \resizebox{0.8\linewidth}{!}{%
    \begin{tabular}{| *{13}{c|}}
        \hline
        \multirow{3}{*}{Model}  &   \multicolumn{6}{c|}{SpaCy Embeddings}   &   \multicolumn{6}{c|}{InferSent Embeddings}   \\
        \cline{2-13}
        &   \multicolumn{2}{c|}{DBI}    &   \multicolumn{2}{c|}{SIL}     &   \multicolumn{2}{c|}{CHS}    & \multicolumn{2}{c|}{DBI} & \multicolumn{2}{c|}{SIL}    & \multicolumn{2}{c|}{CHS} \\
        \cline{2-13}
        & AVG & MIN & AVG & MAX & AVG & MAX & AVG & MIN & AVG & MAX & AVG & MAX \\
        \hline
        Orig.   &   2.72  &	2.72	&   0.10	&   0.10	&   684.14    &	684.32  &  2.85    &	2.85	&   0.092   &	0.092   &	592.55  &	592.77\\
        \textbf{AE}      &   2.18	&   \textbf{0}	    &   0.28  &	\textbf{1}       &	\textbf{207846.94} &	\textbf{1036948.01} &   1.88	&   1.30   &	\textbf{0.68}	&   \textbf{0.99}	&   675.47	&   1315.52    \\
        LC(D)   &   0.71  &	0.59	&   0.55	&   0.59	&   11291.57	&   15142.86    &   0.70	&   0.56	&   0.61  &	0.64    &	11092.32  &	12723.00   \\
        \textbf{AE(D)}   &   \textbf{0.38}  &	0.34   &	\textbf{0.76}  &	0.79    &	24172.28  &	27989.27    &   \textbf{0.60}	&   \textbf{0.52}   &	0.67	&   0.70	&   \textbf{12275.96}	&   \textbf{13767.69}\\
        \hline
    \end{tabular}
    }
    \caption{Using a composite loss, which minimizes binary cross entropy loss and minimizes entanglement through soft nearest neighbor loss, has a significant increase in clustering quality on the AG News dataset using an autoencoder network.}
    \label{tab:internal_criteria_clustering}
\end{table*}
\begin{table*}[h]
    \centering
    \resizebox{0.75\linewidth}{!}{%
    \begin{tabular}{| *{13}{c|}}
        \hline
        \multirow{3}{*}{Model}  &   \multicolumn{6}{c|}{SpaCy Embeddings}   &   \multicolumn{6}{c|}{InferSent Embeddings}   \\
        \cline{2-13}
        &   \multicolumn{2}{c|}{NMI}    &   \multicolumn{2}{c|}{ARI}     &   \multicolumn{2}{c|}{ACC}    & \multicolumn{2}{c|}{NMI} & \multicolumn{2}{c|}{ARI}    & \multicolumn{2}{c|}{ACC} \\
        \cline{2-13}
        & AVG & MAX & AVG & MAX & AVG & MAX & AVG & MAX & AVG & MAX & AVG & MAX \\
        \hline
        Orig.   &   0.51	&   0.51	&   0.52	&   0.52   &	0.78	&   0.79    &   0.53	&   0.53	&   0.57	&   0.57    &	0.81    &	0.81    \\
        AE      &   0.44	&   0.57	&   0.46	&   0.61	&   0.70  &	0.83    &   0.06	&   0.28	&   0.04    &	0.19	&   0.30	&   0.49    \\
        \textbf{LC(D)}   &   0.67	&   \textbf{0.68}	&   0.71	&   \textbf{0.72}	&   \textbf{0.88}	&   0.88    &	0.64	&   0.66    &	0.66	&   0.70    &	\textbf{0.86}    &	\textbf{0.88}    \\
        \textbf{AE(D)}   &   \textbf{0.68}  &	\textbf{0.68}   &	\textbf{0.72}  &	\textbf{0.72}   &	\textbf{0.88}  &	\textbf{0.89}   &   \textbf{0.65}	&   \textbf{0.67}    &	\textbf{0.67}    &	\textbf{0.71}    &	\textbf{0.86}    &	\textbf{0.88}   \\
        \hline
    \end{tabular}}
    \caption{Using a composite loss, which minimizes binary cross entropy loss and minimizes entanglement through soft nearest neighbor loss, has a significant increase in clustering membership performance on the AG News dataset using an autoencoder network.}
    \label{tab:external_criteria_clustering}
\end{table*}

\subsection{Clustering Performance Evaluation}
Since autoencoder and clustering are unsupervised learning techniques, we simulate the lack of labelled datasets by using only a subset of 10,000 labelled examples for training our InferSent models while we still use the 20,000 labelled examples for training our SpaCy models. For our baseline models, we encode the original feature representation (which we denote by ``Orig.'') using principal components analysis to 128 dimensions (the same dimensionality used by our autoencoders), and we use the latent code representation from our baseline autoencoder. The resulting lower-dimensional representation of our sentence embeddings are the inputs to our k-Means clustering model.\\
In Table \ref{tab:internal_criteria_clustering}, we can see that encoding the sentence embeddings using an autoencoder network improves the clustering quality of the learned representations. When using the averaged word embeddings, we can see that our autoencoder with disentangled hidden layers outperformed all the models in terms of DBI and SIL on average over our five runs, but lost significantly in terms of CHS due to one of the runs (seed 31415926) on the baseline autoencoder had 1,036,948.01 as its CHS (and DBI of 0 and SIL of 1). However, if we exclude the baseline autoencoder run on seed 31415926, it has only 571.67 as its CHS. But when using the InferSent embeddings, we see a more consistent result of our experimental models outperforming our baseline models, and only losing to our baseline autoencoder by a slim margin in terms of SIL.\\
In Table \ref{tab:external_criteria_clustering}, we see a significant clustering membership improvement when using disentangled representations for clustering. Surprisingly, even without any ground-truth information unlike our experimental models, the baseline autoencoder performed well when using the averaged word embeddings. Similarly, we have a decent baseline performance as well on the original feature representation for both the averaged word embeddings and InferSent embeddings. For the latter, we cite Cover's theorem which states that by transforming a non-linearly separable data into higher-dimensional space, we can obtain a linearly separable representation of the data, noting that the averaged word embeddings and InferSent embeddings had 300 and 4096 dimensions respectively at the begining, and only transformed to 128 dimensions by using PCA. Furthermore, this even supports the study by \citeauthor{wieting2019no}, where their random encoders performed on par with the models in \cite{conneau2017supervised}, suggesting that the sentence encoding in 4096 dimensions takes the most credit for rendering the classification task easier. As for our experimental models, both disentangling the latent code layer and all the hidden layers of the autoencoder improved the clustering performance in terms of cluster membership as indicated by their high NMI, ARI, and ACC scores.
\begin{figure*}[h]
    \centering
    \includegraphics[width=0.8\linewidth]{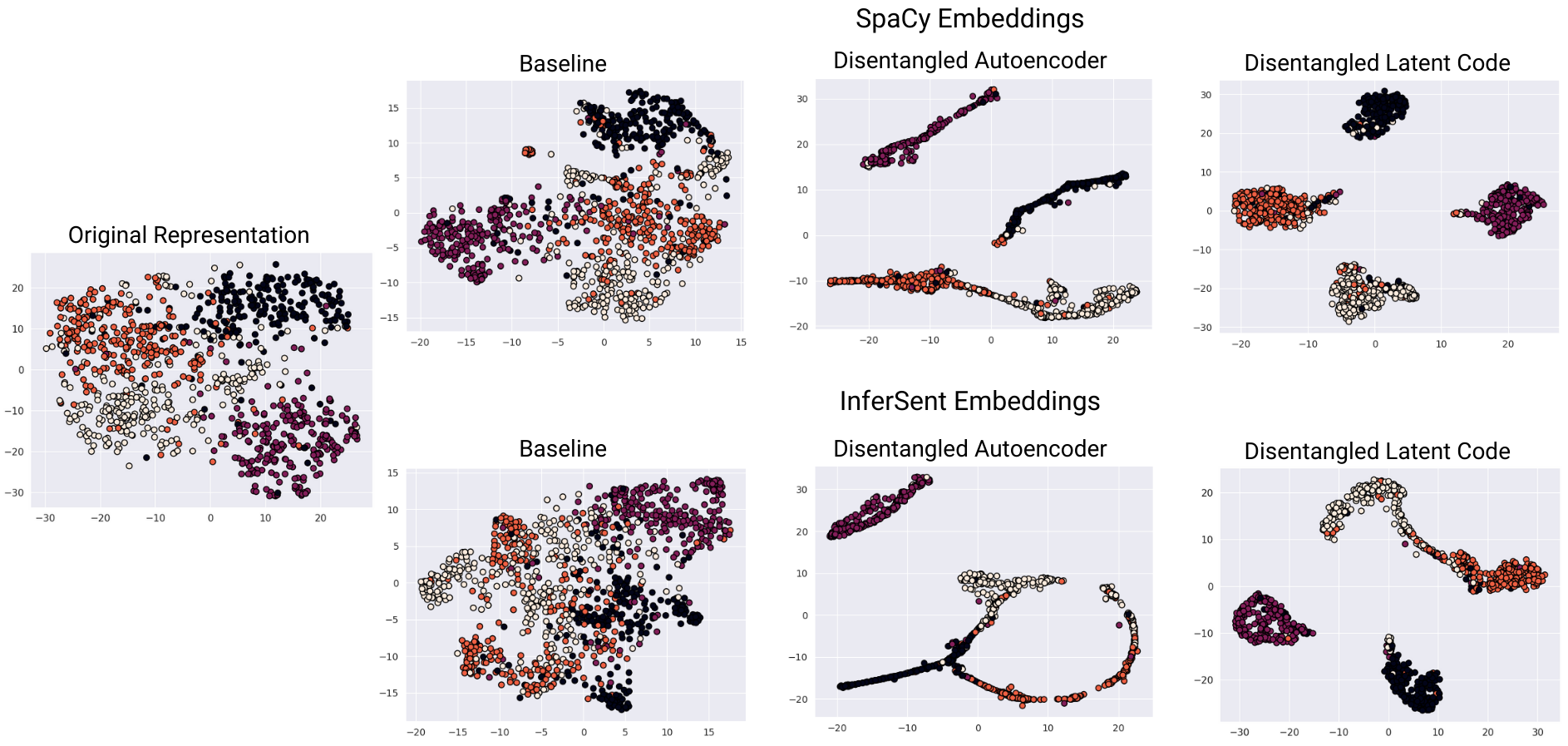}
    \caption{Visualization comparing the original representation with the learned latent code representations by the baseline autoencoder, our autoencoder with disentangled hidden layers, and our autoencoder with disentangled latent code layer. To achieve this visualization, the representations were encoded using t-SNE \cite{maaten2008visualizing} with perplexity = 50.0 and learning rate = 200.0, optimized for 1,000 iterations, using the same random seed for all the computations. This figure is best viewed in color.}
    \label{fig:visualizing_clusters}
\end{figure*}
\subsection{Visualizing Disentangled Text Representations}
We show the latent code representations for both the averaged GloVe embeddings and the InferSent embeddings using our baseline autoencoder, autoencoder with disentangled hidden layers, and autoencoder with disentangled latent code layer in Figure \ref{fig:visualizing_clusters}, together with the original InferSent embeddings representation. We obtain these latent code representations from our clustering experiments. Visually, we can confirm that with disentanglement, we were able to have better-defined clusters.

\section{Conclusion and Future Directions}
To the best of our knowledge, this is the first work on disentangling natural language representations for classification and clustering since the seminal papers on soft nearest neighbor loss focused on image classification and image generation tasks \cite{frosst2019analyzing, goldberger2005neighbourhood, salakhutdinov2007learning}. Our experimental models consistently outperformed our baseline models with marginal improvement on text classification and significant improvement on text clustering, both in the form of sentence embeddings. Moreoever, our findings show that there is only a marginal improvement in performance when using actual sentence embeddings over an averaged word embeddings. We aim to apply disentanglement on more natural language processing tasks such as langauge generation, sentiment analysis, and word embeddings analysis. We also aim to use more datasets for a stronger set of empirical results.

\bibliography{paper}
\bibliographystyle{icml2019}

\end{document}